\documentclass[11pt]{article}

\usepackage[preprint]{acl}

\usepackage{times}
\usepackage{latexsym}
\usepackage{inconsolata}
\usepackage{booktabs}
\usepackage{array}
\usepackage{multirow}
\usepackage{tabularx}

\usepackage[T1]{fontenc}

\usepackage[utf8]{inputenc}

\usepackage{microtype}

\usepackage{graphicx}
\usepackage{xcolor}
\usepackage{tikz}
\usetikzlibrary{arrows.meta,positioning,fit,calc}

\definecolor{covblue}{RGB}{102,157,246}
\definecolor{transcyan}{RGB}{91,192,190}
\definecolor{proxyorange}{RGB}{245,166,90}
\definecolor{ecogreen}{RGB}{112,193,115}
\definecolor{adaptpurple}{RGB}{170,135,242}
\definecolor{softgray}{RGB}{245,245,245}
\definecolor{darktext}{RGB}{50,50,50}

\title{Toward Culturally Grounded Natural Language Processing}

\author{Sina Bagheri Nezhad \\
  Independent Researcher \\
  Seattle, WA, USA \\
  \texttt{sina.bagherinezhad@gmail.com}}

\begin{document}
\maketitle

\begin{abstract}
Multilingual NLP is often treated as a route to global inclusion, but linguistic coverage and cultural competence frequently diverge. This paper synthesizes over 50 papers spanning multilingual performance inequality, cross-lingual transfer, culture-aware evaluation, cultural alignment, multimodal benchmarks, benchmark-design critique, and community-grounded data practices. Across this literature, training data coverage remains important, but outcomes are also shaped by tokenization, prompt language, translated benchmark design, culturally grounded supervision, modality, and who authors or validates evaluation data. We argue that culturally grounded NLP should move beyond treating languages as isolated rows in benchmark tables and instead model communicative ecologies: the institutions, scripts, domains, modalities, and communities through which language is used. We propose a layered evaluation and reporting agenda centered on representation audits, mixed elicitation, ecological validity, community validation, adaptation provenance, within-language variation, and maintenance of living cultural resources.
\end{abstract}

\section{Introduction}

Multilingual NLP is increasingly presented as a route to global inclusion, yet a growing body of work shows that linguistic coverage and cultural competence are not the same thing. Recent surveys note that culture in NLP is multi-scalar, dynamic, and typically operationalized only through partial proxies, while sociocultural critiques argue that many current evaluations treat culture as a static property rather than a situated social process \citep{adilazuarda-etal-2024-towards,liu-etal-2025-culturally,pawar-etal-2025-survey,zhou-etal-2025-culture}. In practice, a model can be fluent in many languages and still mis-handle local norms, misread culturally specific entities, or privilege a dominant worldview in text, multimodal, and interactional settings \citep{rystrom-etal-2025-multilingual,naous-etal-2024-beer,naous-xu-2025-origin,belay-etal-2025-culemo,nayak-etal-2024-benchmarking,havaldar-etal-2025-culturally}.

This concern matters because multilingual NLP has advanced rapidly. Work on global coverage and performance inequality shows that most languages remain weakly represented in resources, benchmarks, and deployed systems \citep{joshi-etal-2020-state,blasi-etal-2022-systematic}. Even when evaluation expands to 200+ languages, large disparities persist \citep{adelani-etal-2024-sib,nllb-team-2022-nllb}. A complementary line of research studies the drivers of cross-lingual transfer, highlighting pretraining distribution, lexical overlap, script, tokenization, and typology as important predictors \citep{philippy-etal-2023-towards,limisiewicz-etal-2023-tokenization,hammerl-etal-2025-beyond,bagheri-nezhad-agrawal-2024-drives,bagheri-nezhad-etal-2025-beyond}. But these predictors alone do not explain whether models are aligned with the values, interactional norms, or situated knowledge of the communities they are meant to serve.

Recent culture-focused benchmarks and adaptation methods make that gap visible. On text evaluation, translated benchmarks can preserve source-culture assumptions, cultural ranking changes appear when evaluation is restricted to culturally sensitive subsets, and survey-style or value-oriented probes reveal substantial cultural dominance in default model behavior \citep{singh-etal-2025-global,wang-etal-2024-cdeval,zhao-etal-2024-worldvaluesbench,alkhamissi-etal-2024-investigating,masoud-etal-2025-cultural,xu-etal-2025-self}. In multimodal and interactional settings, models struggle with culturally grounded VQA, emotion understanding, metaphor interpretation, conversational adaptation, and region-specific question answering \citep{nayak-etal-2024-benchmarking,schneider-etal-2025-gimmick,winata-etal-2025-worldcuisines,maji-etal-2025-drishtikon,yang-etal-2025-cultural,havaldar-etal-2025-culturally}. At the same time, adaptation methods based on native preference data, culture-aware prompting, synthetic critique data, and cultural learning show that these failures are not immutable, but they also reveal how dependent progress is on locally grounded supervision \citep{guo-etal-2025-care,liu-etal-2025-culturally,feng-etal-2025-culfit}. 

We use \emph{communicative ecologies} to refer to the social and technical conditions through which language is produced, interpreted, translated, annotated, evaluated, and deployed, including institutions, scripts, domains, modalities, and communities of practice.

This synthesis integrates evidence from over 50 papers spanning multilingual performance inequality, cross-lingual transfer, culture-aware evaluation, cultural alignment, multimodal benchmarking, benchmark-design critique, and community-grounded data practices. We prioritize recent ACL Anthology, TACL/CL, and C3NLP-adjacent work that either evaluates cultural behavior, explains multilingual performance variation, proposes culture-aware adaptation methods, or documents community-grounded data practices. We make three contributions. First, we connect the multilingual transfer literature to recent culture-oriented evaluation work, showing where the two have often been discussed separately. Second, we identify recurring empirical patterns across the literature: scale matters, but benchmark design, tokenizer behavior, prompt language, local supervision, and modality all materially affect cultural performance. Third, we propose a research agenda for culturally grounded NLP centered on communicative ecologies rather than isolated language labels. Figure~\ref{fig:communicative_ecology} summarizes this shift.

Because the cited literature is broad and recent, our goal is not a formal meta-analysis or an exhaustive survey. Instead, we synthesize representative findings that jointly explain why multilinguality, cultural evaluation, and participatory localization should be studied together.

\begin{figure*}[t]
\centering
\small
\begin{tikzpicture}[
  font=\small,
  card/.style={
    draw,
    rounded corners=4pt,
    thick,
    align=center,
    text width=.145\textwidth,
    minimum height=3.25cm,
    inner sep=6pt,
    text=darktext
  },
  group/.style={
    draw,
    dashed,
    rounded corners=6pt,
    inner sep=8pt,
    thick
  },
  note/.style={
    draw,
    rounded corners=4pt,
    thick,
    align=center,
    text width=.35\textwidth,
    minimum height=1.1cm,
    inner sep=6pt,
    fill=softgray,
    text=darktext
  },
  takeaway/.style={
    draw,
    rounded corners=5pt,
    thick,
    align=center,
    text width=0.97\textwidth,
    minimum height=.95cm,
    inner sep=6pt,
    fill=black!4,
    text=darktext
  },
  arr/.style={-{Latex[length=2.2mm]}, thick},
  node distance=0.32cm and 0.28cm
]

\node[card, fill=covblue!20, draw=covblue!70!black] (c1) {%
\textbf{Multilingual coverage}\\
{\footnotesize languages, benchmarks, resources}

\vspace{1.5mm}
\rule{0.82\linewidth}{0.4pt}

\vspace{1.5mm}
\emph{Coverage question}\\
How many languages or resources are included?
};

\node[card, fill=transcyan!22, draw=transcyan!70!black, right=of c1] (c2) {%
\textbf{Transfer mechanics}\\
{\footnotesize data overlap, script, tokenization}

\vspace{1.5mm}
\rule{0.82\linewidth}{0.4pt}

\vspace{1.5mm}
\emph{Mechanism question}\\
What enables or blocks cross-lingual generalization?
};

\node[card, fill=proxyorange!24, draw=proxyorange!75!black, right=of c2] (c3) {%
\textbf{Cultural proxies}\\
{\footnotesize country labels, surveys, translated prompts}

\vspace{1.5mm}
\rule{0.82\linewidth}{0.4pt}

\vspace{1.5mm}
\emph{Measurement question}\\
Which indirect signals are standing in for culture?
};

\node[card, fill=ecogreen!22, draw=ecogreen!70!black, right=0.8 of c3] (c4) {%
\textbf{Communicative ecologies}\\
{\footnotesize communities, institutions, domains, modalities}

\vspace{1.5mm}
\rule{0.82\linewidth}{0.4pt}

\vspace{1.5mm}
\emph{Context question}\\
Who uses language, where, and under what norms?
};

\node[card, fill=adaptpurple!22, draw=adaptpurple!75!black, right=of c4] (c5) {%
\textbf{Evaluation \& adaptation}\\
{\footnotesize slices, validation, reporting, tuning}

\vspace{1.5mm}
\rule{0.82\linewidth}{0.4pt}

\vspace{1.5mm}
\emph{Action question}\\
How should systems be evaluated, adapted, and maintained?
};

\draw[arr] (c1) -- (c2);
\draw[arr] (c2) -- (c3);
\draw[arr] (c3) -- (c4);
\draw[arr] (c4) -- (c5);

\node[group, draw=black!55, fit=(c1)(c2)(c3),
      label=above:{\textbf{Common multilingual framing}}] (g1) {};

\node[group, draw=black!55, fit=(c4)(c5),
      label=above:{\textbf{Toward culturally grounded NLP}}] (g2) {};

\node[note, below=0.3cm of g1] (n1) {%
\textbf{Risk:} Language-level gains can mask within-language, regional, or community-level failures.
};

\node[note, below=0.3cm of g2] (n2) {%
\textbf{Implication:} Community validation can change both what is measured and what counts as success.
};

\draw[dashed, thick, black!55] (n1.north) -- (g1.south);
\draw[dashed, thick, black!55] (n2.north) -- (g2.south);

\node[takeaway, anchor=north] (take)
  at ([yshift=-0.45cm]$(current bounding box.south west)!0.5!(current bounding box.south east)$) {%
\textbf{Key shift:} from treating \emph{languages as benchmark rows} to modeling \emph{culture as situated practice} within communicative ecologies, supported by community-grounded evaluation and adaptation.
};

\end{tikzpicture}

\caption{From multilingual coverage to culturally grounded NLP. The first three stages reflect common multilingual framing: increasing language coverage, explaining cross-lingual transfer, and approximating culture through indirect proxies. The paper argues for a further shift toward communicative ecologies, where language use is situated in communities, institutions, domains, and modalities, and where evaluation and adaptation are grounded in slice-based analysis, community validation, and explicit reporting.}
\label{fig:communicative_ecology}
\end{figure*}

\begin{table*}[t]
\centering
\small
\setlength{\tabcolsep}{4pt}
\begin{tabular}{p{0.18\textwidth}p{0.46\textwidth}p{0.30\textwidth}}
\toprule
Theme & Representative literature & Main implication \\
\midrule
Coverage and disparity & \citep{joshi-etal-2020-state,blasi-etal-2022-systematic,adelani-etal-2024-sib,nllb-team-2022-nllb} & Aggregate multilingual progress still leaves many languages and communities poorly served. \\
Transfer factors & \citep{philippy-etal-2023-towards,limisiewicz-etal-2023-tokenization,hammerl-etal-2025-beyond,bagheri-nezhad-agrawal-2024-drives,bagheri-nezhad-etal-2025-beyond} & Pretraining, script, and tokenizer behavior shape transfer, but they do not by themselves measure cultural fitness. \\
Culture definitions and surveys & \citep{adilazuarda-etal-2024-towards,liu-etal-2025-culturally,pawar-etal-2025-survey,zhou-etal-2025-culture} & ``Culture'' is multi-faceted and usually operationalized through incomplete proxies. \\
Text and value-oriented evaluation & \citep{wang-etal-2024-cdeval,zhao-etal-2024-worldvaluesbench,singh-etal-2025-global,kim-lee-2025-nunchi,chiu-etal-2025-culturalbench,rystrom-etal-2025-multilingual} & Model rankings and failure modes change on culturally sensitive, locally authored, or culturally situated subsets. \\
Alignment and adaptation & \citep{alkhamissi-etal-2024-investigating,masoud-etal-2025-cultural,xu-etal-2025-self,guo-etal-2025-care,liu-etal-2025-culturally,feng-etal-2025-culfit} & Cultural behavior can shift with prompts or targeted supervision, but gains depend on grounded data and task framing. \\
Multimodal, interactional, and local tasks & \citep{nayak-etal-2024-benchmarking,schneider-etal-2025-gimmick,yang-etal-2025-cultural,belay-etal-2025-culemo,maji-etal-2025-drishtikon,winata-etal-2025-worldcuisines,villa-cueva-etal-2025-cammt,nyandwi-etal-2025-grounding,havaldar-etal-2025-culturally,ochieng-etal-2025-beyond,aji-cohn-2025-loraxbench,ferawati-etal-2024-synchronizing,pranida-etal-2025-culturally,naous-etal-2024-beer,naous-xu-2025-origin} & Local knowledge, interactional norms, region-specific entities, and within-language variation remain difficult across settings and modalities. \\
\bottomrule
\end{tabular}
\caption{Representative lines of work that jointly motivate a culture-grounded view of multilingual NLP.}
\label{tab:landscape}
\end{table*}

\section{From Multilinguality to Cultural Competence}

\subsection{Coverage, transfer, and the limits of language-only evaluation}

The multilingual literature has established two robust facts. First, language technology is unequally distributed. Resource concentration, benchmark availability, and deployment all remain heavily skewed toward a small subset of languages \citep{joshi-etal-2020-state,blasi-etal-2022-systematic}. Second, widening language coverage does not remove disparity by itself. When evaluation is extended to hundreds of languages and dialects, performance gaps remain large and often correlate with resource concentration or infrastructure advantages \citep{adelani-etal-2024-sib,nllb-team-2022-nllb,aji-cohn-2025-loraxbench}.

Research on cross-lingual transfer helps explain part of this pattern. Reviews and empirical analyses point to pretraining data, lexical sharing, script, tokenization, and other similarity signals as important determinants of multilingual transfer \citep{philippy-etal-2023-towards,limisiewicz-etal-2023-tokenization,hammerl-etal-2025-beyond,bagheri-nezhad-agrawal-2024-drives,bagheri-nezhad-etal-2025-beyond}. These findings are valuable, but they still mostly answer a language-centric question: \emph{when does one language help another?} They say much less about whether the model understands local norms, values, or region-specific knowledge once it has transferred technically.

That distinction matters because token or script overlap can support transfer while still leaving culturally grounded behavior under-specified. A model may reuse familiar subwords and achieve respectable benchmark accuracy, yet still fail when asked to interpret a culturally loaded metaphor, a local emotional expression, a region-specific entity, or a social norm that is not well represented in pretraining data \citep{naous-etal-2024-beer,naous-xu-2025-origin,belay-etal-2025-culemo,yang-etal-2025-cultural,nyandwi-etal-2025-grounding}. Multilinguality, then, is necessary but not sufficient for cultural competence.

\subsection{Culture is usually operationalized through proxies}

A second lesson from recent surveys is conceptual. ``Culture'' is rarely modeled directly; instead, it is approximated through countries, languages, prompt languages, values surveys, food, rituals, emotions, conversational style, or locally salient entities \citep{adilazuarda-etal-2024-towards,liu-etal-2025-culturally,pawar-etal-2025-survey}. This is unavoidable to some extent, but it also creates risk. A benchmark may appear ``cross-cultural'' while actually testing only one thin slice of culture, such as national-value questionnaires or culture-specific trivia.

Recent conceptual work argues that this is not just a measurement inconvenience but a theoretical problem. \citet{zhou-etal-2025-culture} show that many cultural NLP setups rely on static and often nationalized proxies, while \citet{liu-etal-2025-culturally} emphasize that culture spans within-human, between-human, and extra-human dimensions. In related work, alignment studies operationalize culture through prompt language, value surveys, or persona framing and demonstrate that model behavior can shift under these interventions \citep{alkhamissi-etal-2024-investigating,masoud-etal-2025-cultural,xu-etal-2025-self}. The implication is not that such proxies are useless, but that they should be read as partial windows into broader communicative ecologies rather than as full representations of culture.

\section{What Recent Benchmarks and Adaptation Work Show}

\subsection{Text and value-oriented evaluation}

Recent text-centric evaluation work makes the multilingual-versus-multicultural gap especially clear. Global-MMLU shows that translated benchmarks can preserve linguistic and cultural assumptions from English source questions, and that model rankings may shift when evaluation is restricted to culturally sensitive subsets \citep{singh-etal-2025-global}. Related benchmarks such as CDEval and WorldValuesBench probe cultural dimensions or value distributions more directly, revealing that models can produce systematically different alignments across countries and social domains \citep{wang-etal-2024-cdeval,zhao-etal-2024-worldvaluesbench}. CulturalBench and Nunchi-Bench further show that locally grounded, human-written questions expose failures that are hard to detect with standard multilingual QA or classification tasks \citep{chiu-etal-2025-culturalbench,kim-lee-2025-nunchi}.

This gap is also visible in work on cultural bias. In Arabic, \citet{naous-etal-2024-beer} show that multilingual and monolingual models can prefer Western-associated entities and generate culturally inappropriate associations. \citet{naous-xu-2025-origin} trace part of this bias to the interaction between pretraining distributions, linguistic phenomena, and frequency-based tokenization, illustrating that technical design choices can have culturally asymmetric effects. More broadly, \citet{rystrom-etal-2025-multilingual} find that multilingual ability does not reliably predict cultural alignment, reinforcing the argument that language coverage and cultural competence should be evaluated separately.

Text evaluation also reveals that alignment can move, but unevenly. Prompt language and persona prompting affect cultural outputs in both value-oriented and survey-based setups \citep{alkhamissi-etal-2024-investigating,masoud-etal-2025-cultural}. Self-pluralising or culture-joint tuning can improve alignment to multiple cultures without fully collapsing general performance, but the underlying problem remains open because cultures are internally diverse and task demands differ \citep{xu-etal-2025-self}. The shared lesson is that cultural performance is not a fixed model property; it is partly a function of how the task is framed, authored, and prompted.

\subsection{Multimodal, interactional, and local evaluation}

Culture is not expressed only in text, and recent multimodal work makes this plain. CulturalVQA, GIMMICK, WorldCuisines, DRISHTIKON, and CaMMT all show that models struggle with geographically diverse food, dress, artifacts, rituals, and region-specific visual cues \citep{nayak-etal-2024-benchmarking,schneider-etal-2025-gimmick,winata-etal-2025-worldcuisines,maji-etal-2025-drishtikon,villa-cueva-etal-2025-cammt}. Work on multimodal metaphors and culturally grounded MLLMs reaches a similar conclusion: local entities and cultural references remain brittle, especially when they are rare in dominant training sources \citep{yang-etal-2025-cultural,nyandwi-etal-2025-grounding}.

Interactional and low-resource settings reveal additional failure modes. CULEMO shows that cross-cultural emotion understanding varies substantially across languages and resource conditions \citep{belay-etal-2025-culemo}. Culturally-Aware Conversations argues that many existing benchmarks are misaligned with the actual conversational situations in which cultural adaptation matters, and demonstrates that even strong models struggle with stylistic sensitivity and subjective correctness \citep{havaldar-etal-2025-culturally}. In low-resource real-world scenarios, qualitative assessment can surface weaknesses that automatic metrics miss, especially in code-mixed or socially contextualized tasks \citep{ochieng-etal-2025-beyond}. Regionally focused benchmarks such as LORAXBENCH and culturally nuanced story-generation tasks for Javanese and Sundanese also show that cultural reasoning, formality, and local narrative competence are uneven even within one geopolitical region \citep{aji-cohn-2025-loraxbench,pranida-etal-2025-culturally}.

These findings also connect back to dataset construction. Synchronizing annotation guidelines across cultures is itself a methodological challenge, as shown by work on multilingual annotation design \citep{ferawati-etal-2024-synchronizing}. In other words, cultural competence is not only something to test at the end of the pipeline; it is also shaped by how annotation, adjudication, and benchmark authorship are organized from the start.

\subsection{Adaptation and alignment methods}

Recent alignment work suggests that targeted intervention can help. Early cultural-alignment studies showed that model responses shift when prompted with a dominant local language or culture-specific persona information \citep{alkhamissi-etal-2024-investigating}. More recent methods go further by introducing culture-aware training signals. CARE uses multilingual human preference data from native speakers to improve cultural awareness \citep{guo-etal-2025-care}; CLCA uses simulated social interactions to adapt models toward target cultural values \citep{liu-etal-2025-cultural}; CulFiT synthesizes multilingual critiques to support fine-grained cultural training \citep{feng-etal-2025-culfit}; and CultureSPA demonstrates that pluralistic culture alignment can be improved through self-generated culture-related supervision \citep{xu-etal-2025-self}.

These methods are promising, but their success underscores a broader point: cultural competence does not arise automatically from multilingual scale. It often requires native raters, culture-specific task framing, or carefully designed supervision \citep{guo-etal-2025-care,havaldar-etal-2025-culturally}. The data and design burden is therefore part of the science, not just an implementation detail.

\subsection{Evaluation paradigms, representation, and ecological validity}

Recent work increasingly argues that the main bottleneck is not only model capability but also what current evaluations choose to treat as culture. \citet{oh-etal-2025-culture} describe a ``trivia-centered'' paradigm in which culture is reduced to static facts, survey answers, or decontextualized commonsense questions. \citet{kabir-etal-2025-break} show that this matters empirically: open-ended assessment can yield different conclusions from closed multiple-choice evaluation, because it lets models express partial knowledge, uncertainty, or culturally plausible alternatives that a fixed answer set suppresses. Meanwhile, \citet{wu-etal-2025-incorporating} survey 105 cultural-alignment benchmarks and find that representation is heavily concentrated in majority populations, dominant languages, and national-level categories, with much less attention to subcultural variation or minority perspectives. \citet{alkhamissi-etal-2026-hire} extend the critique by arguing that many benchmarks rely on impoverished concepts of culture and would benefit from stronger anthropological grounding.

These measurement questions are tightly linked to task realism. Cultural failures often appear most clearly when a model must act in a situated environment rather than answer a standalone prompt. \citet{qiu-etal-2025-evaluating} show that LLM web agents have markedly weaker cultural and social awareness in live browsing settings than in simpler, non-agentic setups. Likewise, recent multimodal video benchmarks indicate that cultural competence depends on joint interpretation of scene context, gesture, speech, objects, and event structure, not just isolated text or static images \citep{shafique-etal-2025-culturally}. Ecological validity is therefore not a downstream deployment issue; it is part of the evaluation construct itself.

Native-speaker and community-validated datasets sharpen the point further. DaKultur shows that automatically translated data are inadequate for evaluating Danish cultural awareness and that native-speaker data materially improve both human acceptance and the usefulness of automatic evaluation \citep{muller-eberstein-etal-2025-dakultur}. Community-engaged resources such as HESEIA and SAFARI similarly demonstrate that co-design changes which stereotypes, harms, and everyday contexts are even represented in the benchmark \citep{ivetta-etal-2025-heseia,verma-etal-2026-safari}. Representation, in other words, is not only a matter of \emph{how many} languages or regions appear, but of \emph{whose categories, harms, and communicative practices} define the task.

Finally, recent efforts to scale culture-related resources reveal a tension between breadth and grounding. CultureInstruct and newer work on scaling cultural resources suggest that larger multicultural instruction collections can improve performance on cultural benchmarks, yet they also raise familiar questions about provenance, normalization, and loss of local detail \citep{pham-etal-2025-cultureinstruct,stepanyan-etal-2026-scaling}. The research challenge is therefore not just to collect more cultural data, but to preserve enough social context that scaling does not erase the very phenomena the field claims to model.

\section{Synthesis: Recurring Findings Across the Literature}

\subsection{Data quantity remains necessary but insufficient}

Across the multilingual literature, training data coverage remains one of the strongest predictors of performance \citep{joshi-etal-2020-state,blasi-etal-2022-systematic,adelani-etal-2024-sib,bagheri-nezhad-agrawal-2024-drives,bagheri-nezhad-etal-2025-beyond}. However, recent work shows that the path from coverage to performance is mediated by tokenizer allocation, literal and non-literal token alignability, prompt language, and the interaction between local linguistic phenomena and pretraining composition \citep{limisiewicz-etal-2023-tokenization,hammerl-etal-2025-beyond,naous-xu-2025-origin}. More data is therefore necessary, but it does not guarantee that a model will reason appropriately about local culture once deployed.

\subsection{Benchmark design often imports external cultural assumptions}

A second recurring pattern is that benchmark construction itself can flatten culture. Translation from English source tasks can import assumptions about curricula, answer spaces, or what counts as common knowledge \citep{singh-etal-2025-global}. Survey-style probes can capture useful signals, but they also narrow culture to the specific dimensions encoded in the survey instrument \citep{wang-etal-2024-cdeval,zhao-etal-2024-worldvaluesbench,masoud-etal-2025-cultural}. Even locally grounded benchmarks face design choices around question writing, annotation, and disagreement handling \citep{chiu-etal-2025-culturalbench,kim-lee-2025-nunchi,havaldar-etal-2025-culturally,ferawati-etal-2024-synchronizing}. The implication is that culture-sensitive evaluation must report not only scores, but also the assumptions embedded in benchmark construction.

\subsection{Local supervision and participation matter}

A third pattern is methodological. When native speakers, culturally diverse raters, or locally grounded supervision are included, performance and diagnostic value tend to improve \citep{guo-etal-2025-care,liu-etal-2025-culturally,feng-etal-2025-culfit,havaldar-etal-2025-culturally}. Conversely, low-resource studies repeatedly show that generic metrics or imported annotation schemes can obscure the very failures that matter in practice \citep{ochieng-etal-2025-beyond,aji-cohn-2025-loraxbench}. Participation is therefore not only an ethical add-on; it changes what the field can validly claim to have measured.

\subsection{Culture is distributed across modalities and within-language variation}

The broadest lesson from recent work is that culture is distributed. It appears in images, food, artifacts, emotional repertoires, social interaction, metaphors, politeness registers, and local entities, not only in standardized written text \citep{nayak-etal-2024-benchmarking,schneider-etal-2025-gimmick,yang-etal-2025-cultural,belay-etal-2025-culemo,maji-etal-2025-drishtikon,winata-etal-2025-worldcuisines,villa-cueva-etal-2025-cammt,nyandwi-etal-2025-grounding}. It also appears within languages, through dialect, register, code-mixing, and local narrative conventions \citep{aji-cohn-2025-loraxbench,pranida-etal-2025-culturally,ochieng-etal-2025-beyond}. This means that culture-sensitive NLP cannot be reduced to a one-language-one-culture mapping.

\subsection{From language lists to communicative ecologies}

Taken together, these findings suggest a shift in framing. Languages should not be treated as isolated rows in a benchmark spreadsheet, but as elements of communicative ecologies: institutions, scripts, education systems, media networks, annotation pipelines, and communities of practice. Some multilingual transfer studies already hint at this by showing that contextual covariates beyond raw data amount can help explain performance \citep{bagheri-nezhad-etal-2025-beyond}. Culture-oriented work extends the point by showing that what matters is not only \emph{which language} is present, but \emph{how that language is socially situated} in training, evaluation, and deployment \citep{liu-etal-2025-culturally,zhou-etal-2025-culture,rystrom-etal-2025-multilingual}.

\section{Toward Better Evaluation Protocols}

The literature now supports a practical conclusion: evaluating cultural competence should be treated as a layered protocol rather than a single benchmark score. Table~\ref{tab:protocol} summarizes concrete shortcuts and stronger alternatives. The key idea is to separate questions that are often collapsed in current practice: who is represented, how the task elicits cultural knowledge, how realistic the evaluation setting is, who validates outputs, and what kinds of data changed the model.

\begin{table*}[t]
\centering
\small
\setlength{\tabcolsep}{4pt}
\begin{tabular}{p{0.17\textwidth}p{0.22\textwidth}p{0.27\textwidth}p{0.24\textwidth}}
\toprule
Protocol layer & Common shortcut & Stronger practice & Representative literature \\
\midrule
Representation audit & Country/language labels only & Report authorship, language variety, translation pipeline, and whether majority/minority or subcultural groups are represented & \citep{wu-etal-2025-incorporating,alkhamissi-etal-2026-hire,faisal-etal-2024-dialectbench} \\
Elicitation diversity & Multiple-choice or Likert only & Combine closed items with open generation, pairwise judgments, and qualitative error analysis & \citep{oh-etal-2025-culture,kabir-etal-2025-break,havaldar-etal-2025-culturally} \\
Ecological validity & Static text-only QA & Add conversation, web-agent, image, video, and region-specific task slices & \citep{qiu-etal-2025-evaluating,shafique-etal-2025-culturally,nayak-etal-2024-benchmarking} \\
Community validation & Expert-only or automatic scoring & Include native-speaker review, disagreement analysis, and participatory co-design & \citep{muller-eberstein-etal-2025-dakultur,ivetta-etal-2025-heseia,verma-etal-2026-safari} \\
Adaptation reporting & ``Culture-tuned'' as a black box & Publish supervision provenance, target population, and trade-offs across groups and tasks & \citep{guo-etal-2025-care,pham-etal-2025-cultureinstruct,stepanyan-etal-2026-scaling} \\
\bottomrule
\end{tabular}
\caption{A practical protocol for evaluating cultural competence beyond language-only leaderboards.}
\label{tab:protocol}
\end{table*}

Representation audits should accompany every benchmark. At minimum, papers should report whether items are translated or native-authored, which language varieties are included, whether prompts target dominant or minoritized groups, and how local validation was handled. This matters because country and language labels can hide major within-language gaps, as shown by DIALECTBENCH and low-resource narrative tasks \citep{faisal-etal-2024-dialectbench,aji-cohn-2025-loraxbench,ochieng-etal-2025-beyond}. Benchmark audits further suggest that majority-focused scope is still the norm, so missing-group analysis should be treated as core evaluation rather than appendix material \citep{wu-etal-2025-incorporating,alkhamissi-etal-2026-hire}.

Elicitation should also be methodologically diverse. Closed-form surveys and multiple-choice QA remain useful for comparability, but they should be complemented with open generation, pairwise preference judgments, disagreement analysis, and qualitative error coding. Recent work demonstrates that different elicitation styles surface different cultural behaviors and that static trivia-style formats are too narrow for many deployment settings \citep{oh-etal-2025-culture,kabir-etal-2025-break,havaldar-etal-2025-culturally}. Mixed protocols can better distinguish absence of knowledge, culturally plausible variation, and direct norm violation.

A third requirement is to evaluate situated use. This includes multimodal tasks, dialect and code-mixed inputs, region-specific entities, conversation, and agentic interaction. The point is not that every paper must evaluate every modality, but that claims about cultural competence should be limited to the settings actually tested. Results from CulturalVQA, WorldCuisines, video benchmarks, and web agents show that performance can degrade sharply when models must integrate local cues in more realistic contexts \citep{nayak-etal-2024-benchmarking,winata-etal-2025-worldcuisines,shafique-etal-2025-culturally,qiu-etal-2025-evaluating}.

Finally, adaptation papers should document the social provenance of their supervision. When models improve through native preference data, co-designed datasets, or large-scale culture-specific instructions, those data are part of the causal explanation and should be reported as such. Participatory work in low-resource MT and design-inspired NLP already offers a vocabulary for this, while newer datasets show how community involvement changes both task design and measured harms \citep{nekoto-etal-2020-participatory,caselli-etal-2021-guiding,guo-etal-2025-care,ivetta-etal-2025-heseia,verma-etal-2026-safari}. A culture-aware benchmark without provenance or stakeholder documentation may still be useful, but its claims should be correspondingly narrower.

A final implication is temporal maintenance. Cultural resources should be versioned, refreshed, and revalidated as social norms, salient entities, and public events change. Static benchmark release is especially brittle for culturally grounded tasks because local salience can shift faster than generic linguistic competence, and because new communities or subcultures may become visible only after deployment. Recent benchmark-audit and resource-scaling work suggests that ongoing collection and community review are necessary if culture-aware evaluation is to remain current rather than freeze one historical snapshot \citep{wu-etal-2025-incorporating,stepanyan-etal-2026-scaling,verma-etal-2026-safari,ivetta-etal-2025-heseia}.

\section{A Research Agenda for Culturally Grounded NLP}

\paragraph{Distinguish multilingual coverage from cultural competence.} Reporting aggregate multilingual accuracy is not enough. Evaluations should explicitly separate language coverage from culture-sensitive performance, for example by contrasting translated versus native-authored items, culturally sensitive versus culturally agnostic subsets, and monolingual versus code-mixed or interactional settings \citep{singh-etal-2025-global,rystrom-etal-2025-multilingual,ochieng-etal-2025-beyond}.

\paragraph{Publish richer contextual metadata and benchmark slices.} Benchmarks should document script practices, domain concentration, register, source authorship, translation pipeline, represented communities, and community validation. The goal is not to create one scalar ``culture score,'' but to make the assumptions behind evaluation auditable \citep{liu-etal-2025-culturally,zhou-etal-2025-culture,wu-etal-2025-incorporating,alkhamissi-etal-2026-hire}.

\paragraph{Use mixed elicitation protocols.} Closed-form questionnaires are useful for comparability, but culture-sensitive evaluation should combine closed and open-ended elicitation, pairwise judgments, and qualitative analysis, while distinguishing genuine norm violation from reasonable local variation \citep{oh-etal-2025-culture,kabir-etal-2025-break,qiu-etal-2025-evaluating}.

\paragraph{Prefer native-authored and locally validated data where possible.} Recent work shows that native preference data, culturally diverse raters, locally authored prompts, and community-grounded datasets surface different errors and can improve adaptation quality \citep{guo-etal-2025-care,chiu-etal-2025-culturalbench,kim-lee-2025-nunchi,muller-eberstein-etal-2025-dakultur,ivetta-etal-2025-heseia}. This suggests that participation should be treated as core research infrastructure rather than optional review.

\paragraph{Evaluate multimodality and within-language variation.} Future benchmarks should move beyond text-only settings and beyond standardized language varieties. Multimodal artifacts, local entities, politeness registers, dialects, and narrative traditions are central to real deployment contexts \citep{nayak-etal-2024-benchmarking,winata-etal-2025-worldcuisines,maji-etal-2025-drishtikon,aji-cohn-2025-loraxbench,pranida-etal-2025-culturally,faisal-etal-2024-dialectbench,shafique-etal-2025-culturally}.

\paragraph{Treat alignment as ongoing localization rather than one-time correction.} Cultural adaptation is likely to remain task-specific, community-specific, and dynamic. Promising methods already exist, but they should be understood as part of a continuous localization process that must accommodate pluralism and change over time \citep{alkhamissi-etal-2024-investigating,liu-etal-2025-culturally,feng-etal-2025-culfit,xu-etal-2025-self,pham-etal-2025-cultureinstruct,stepanyan-etal-2026-scaling}.

\paragraph{Make participation and governance part of evaluation.} Community-engaged data creation and participatory design should be treated as part of the experimental setup, not just project framing. Earlier work in low-resource MT and positive-impact NLP already provides methodological guidance, and newer culturally grounded datasets show why this matters for which harms and categories are surfaced \citep{nekoto-etal-2020-participatory,caselli-etal-2021-guiding,ivetta-etal-2025-heseia,verma-etal-2026-safari}.

\paragraph{Maintain cultural resources as living infrastructure.} Cultural knowledge, norms, and harms are dynamic. Benchmarks should therefore support refresh cycles, re-annotation, and versioned slices rather than one-off releases that implicitly fix culture at collection time. Audit and scaling studies suggest that this ongoing maintenance is necessary both for coverage and for preventing majority defaults from becoming entrenched as de facto standards \citep{wu-etal-2025-incorporating,alkhamissi-etal-2026-hire,stepanyan-etal-2026-scaling,verma-etal-2026-safari}.

\section{Conclusion}

The central lesson of recent work is straightforward: multilingual capability does not automatically yield multicultural competence. The field now has substantial evidence, across text, values, emotion, conversation, multimodality, and low-resource evaluation, that models can transfer technically while still failing culturally \citep{rystrom-etal-2025-multilingual,singh-etal-2025-global,belay-etal-2025-culemo,schneider-etal-2025-gimmick}. Building more culturally grounded NLP systems will require more than scaling data. It will require better benchmark design, richer contextual metadata, community-grounded supervision, and evaluation protocols that recognize languages as part of larger communicative ecologies.

For the ACL/C3NLP community, the immediate implication is methodological rather than purely rhetorical. Culture-sensitive NLP should become a standard reporting axis alongside multilingual accuracy, robustness, and safety. Papers that claim broad generalization should specify whether they tested translated or native-authored items, standard or non-standard varieties, static or interactive settings, and which communities validated the data and labels. Without that shift, the field risks continuing to reward models that are globally legible yet locally brittle. With it, multilingual NLP can move closer to systems that are not merely widespread, but socially usable across the communities they reach.

\section{Limitations}

This is a synthesis paper rather than a new experimental study, so its claims depend on the coverage and quality of the underlying literature. Although the evidence base spans more than 50 papers, it is still uneven across regions, modalities, and tasks. Many benchmarks are recent, and some target a small number of countries or languages even when their conceptual claims are broad.

A second limitation is that the literature itself often operationalizes culture through partial proxies such as countries, prompt languages, survey instruments, or region-specific artifacts. We follow the literature in discussing these measures, but we do not claim that any one of them captures culture exhaustively. Our use of the term \emph{communicative ecologies} is intended precisely to resist over-identifying culture with any single variable.

A third limitation is temporal. Culture-aware evaluation and alignment are moving quickly, particularly in workshop, multimodal, and low-resource settings. The research agenda proposed here should therefore be read as a living program rather than a closed taxonomy.

\section{Ethical Considerations}

The main ethical risk in operationalizing culture for NLP is reification. Country, language, script, or survey averages can be analytically useful while still flattening internal diversity, especially for Indigenous, diasporic, minoritized, or non-standard language communities \citep{zhou-etal-2025-culture,liu-etal-2025-culturally}. We therefore recommend treating such variables as partial descriptors rather than as direct stand-ins for culture.

A second risk concerns optimization targets. If model development focuses only on predictors that maximize average performance, it may reinforce existing inequalities by favoring languages and communities that already benefit from strong representation, stable orthographies, or abundant culturally relevant supervision \citep{joshi-etal-2020-state,blasi-etal-2022-systematic}. Culture-sensitive evaluation should thus prioritize disparity reporting, documentation, and community relevance in addition to mean scores.

Finally, participatory alignment is not a magic solution. Communities are internally diverse, disagreement is often substantive, and cultural norms change over time. Native preference learning and local annotation should therefore be paired with transparency about whose judgments are represented and what disagreements remain unresolved \citep{guo-etal-2025-care,havaldar-etal-2025-culturally,ferawati-etal-2024-synchronizing}.

\bibliography{c3nlp_acl2026_paper_v4}

@inproceedings{adelani-etal-2024-sib,
    title = "{SIB}-200: A Simple, Inclusive, and Big Evaluation Dataset for Topic Classification in 200+ Languages and Dialects",
    author = "Adelani, David Ifeoluwa  and
      Liu, Hannah  and
      Shen, Xiaoyu  and
      Vassilyev, Nikita  and
      Alabi, Jesujoba O.  and
      Mao, Yanke  and
      Gao, Haonan  and
      Lee, En-Shiun Annie",
    editor = "Graham, Yvette  and
      Purver, Matthew",
    booktitle = "Proceedings of the 18th Conference of the European Chapter of the Association for Computational Linguistics (Volume 1: Long Papers)",
    month = mar,
    year = "2024",
    address = "St. Julian{'}s, Malta",
    publisher = "Association for Computational Linguistics",
    url = "https://aclanthology.org/2024.eacl-long.14/",
    doi = "10.18653/v1/2024.eacl-long.14",
    pages = "226--245"
}

@inproceedings{bagheri-nezhad-agrawal-2024-drives,
    title = "What Drives Performance in Multilingual Language Models?",
    author = "Bagheri Nezhad, Sina  and
      Agrawal, Ameeta",
    editor = {Scherrer, Yves  and
      Jauhiainen, Tommi  and
      Ljube{\v{s}}i{\'c}, Nikola  and
      Zampieri, Marcos  and
      Nakov, Preslav  and
      Tiedemann, J{\"o}rg},
    booktitle = "Proceedings of the Eleventh Workshop on NLP for Similar Languages, Varieties, and Dialects (VarDial 2024)",
    month = jun,
    year = "2024",
    address = "Mexico City, Mexico",
    publisher = "Association for Computational Linguistics",
    url = "https://aclanthology.org/2024.vardial-1.2/",
    doi = "10.18653/v1/2024.vardial-1.2",
    pages = "16--27"
}

@inproceedings{bagheri-nezhad-etal-2025-beyond,
    title = "Beyond Data Quantity: Key Factors Driving Performance in Multilingual Language Models",
    author = "Bagheri Nezhad, Sina  and
      Agrawal, Ameeta  and
      Pokharel, Rhitabrat",
    editor = "Hettiarachchi, Hansi  and
      Ranasinghe, Tharindu  and
      Rayson, Paul  and
      Mitkov, Ruslan  and
      Gaber, Mohamed  and
      Premasiri, Damith  and
      Tan, Fiona Anting  and
      Uyangodage, Lasitha",
    booktitle = "Proceedings of the First Workshop on Language Models for Low-Resource Languages",
    month = jan,
    year = "2025",
    address = "Abu Dhabi, United Arab Emirates",
    publisher = "Association for Computational Linguistics",
    url = "https://aclanthology.org/2025.loreslm-1.18/",
    pages = "225--239"
}

@inproceedings{belay-etal-2025-culemo,
    title = "{CULEMO}: Cultural Lenses on Emotion - Benchmarking {LLM}s for Cross-Cultural Emotion Understanding",
    author = "Belay, Tadesse Destaw  and
      Ahmed, Ahmed Haj  and
      Grissom II, Alvin  and
      Ameer, Iqra  and
      Sidorov, Grigori  and
      Kolesnikova, Olga  and
      Yimam, Seid Muhie",
    editor = "Che, Wanxiang  and
      Nabende, Joyce  and
      Shutova, Ekaterina  and
      Pilehvar, Mohammad Taher",
    booktitle = "Proceedings of the 63rd Annual Meeting of the Association for Computational Linguistics (Volume 1: Long Papers)",
    month = jul,
    year = "2025",
    address = "Vienna, Austria",
    publisher = "Association for Computational Linguistics",
    url = "https://aclanthology.org/2025.acl-long.925/",
    doi = "10.18653/v1/2025.acl-long.925",
    pages = "18894--18909",
    ISBN = "979-8-89176-251-0"
}

@inproceedings{blasi-etal-2022-systematic,
    title = "Systematic Inequalities in Language Technology Performance across the World{'}s Languages",
    author = "Blasi, Damian  and
      Anastasopoulos, Antonios  and
      Neubig, Graham",
    editor = "Muresan, Smaranda  and
      Nakov, Preslav  and
      Villavicencio, Aline",
    booktitle = "Proceedings of the 60th Annual Meeting of the Association for Computational Linguistics (Volume 1: Long Papers)",
    month = may,
    year = "2022",
    address = "Dublin, Ireland",
    publisher = "Association for Computational Linguistics",
    url = "https://aclanthology.org/2022.acl-long.376/",
    doi = "10.18653/v1/2022.acl-long.376",
    pages = "5486--5505"
}

@inproceedings{ferawati-etal-2024-synchronizing,
    title = "Synchronizing Approach in Designing Annotation Guidelines for Multilingual Datasets: A {COVID}-19 Case Study Using {E}nglish and {J}apanese Tweets",
    author = "Ferawati, Kiki  and
      She, Wan Jou  and
      Wakamiya, Shoko  and
      Aramaki, Eiji",
    editor = "Prabhakaran, Vinodkumar  and
      Dev, Sunipa  and
      Benotti, Luciana  and
      Hershcovich, Daniel  and
      Cabello, Laura  and
      Cao, Yong  and
      Adebara, Ife  and
      Zhou, Li",
    booktitle = "Proceedings of the 2nd Workshop on Cross-Cultural Considerations in NLP",
    month = aug,
    year = "2024",
    address = "Bangkok, Thailand",
    publisher = "Association for Computational Linguistics",
    url = "https://aclanthology.org/2024.c3nlp-1.3/",
    doi = "10.18653/v1/2024.c3nlp-1.3",
    pages = "32--41"
}

@inproceedings{feng-etal-2025-culfit,
    title = "{C}ul{F}i{T}: A Fine-grained Cultural-aware {LLM} Training Paradigm via Multilingual Critique Data Synthesis",
    author = "Feng, Ruixiang  and
      Gao, Shen  and
      Chen, Xiuying  and
      Chen, Lisi  and
      Shang, Shuo",
    editor = "Che, Wanxiang  and
      Nabende, Joyce  and
      Shutova, Ekaterina  and
      Pilehvar, Mohammad Taher",
    booktitle = "Proceedings of the 63rd Annual Meeting of the Association for Computational Linguistics (Volume 1: Long Papers)",
    month = jul,
    year = "2025",
    address = "Vienna, Austria",
    publisher = "Association for Computational Linguistics",
    url = "https://aclanthology.org/2025.acl-long.1092/",
    doi = "10.18653/v1/2025.acl-long.1092",
    pages = "22413--22430",
    ISBN = "979-8-89176-251-0"
}

@inproceedings{guo-etal-2025-care,
    title = "{CARE}: Multilingual Human Preference Learning for Cultural Awareness",
    author = "Guo, Geyang  and
      Naous, Tarek  and
      Wakaki, Hiromi  and
      Nishimura, Yukiko  and
      Mitsufuji, Yuki  and
      Ritter, Alan  and
      Xu, Wei",
    editor = "Christodoulopoulos, Christos  and
      Chakraborty, Tanmoy  and
      Rose, Carolyn  and
      Peng, Violet",
    booktitle = "Proceedings of the 2025 Conference on Empirical Methods in Natural Language Processing",
    month = nov,
    year = "2025",
    address = "Suzhou, China",
    publisher = "Association for Computational Linguistics",
    url = "https://aclanthology.org/2025.emnlp-main.1669/",
    doi = "10.18653/v1/2025.emnlp-main.1669",
    pages = "32866--32895",
    ISBN = "979-8-89176-332-6"
}

@inproceedings{joshi-etal-2020-state,
    title = "The State and Fate of Linguistic Diversity and Inclusion in the {NLP} World",
    author = "Joshi, Pratik  and
      Santy, Sebastin  and
      Budhiraja, Amar  and
      Bali, Kalika  and
      Choudhury, Monojit",
    editor = "Jurafsky, Dan  and
      Chai, Joyce  and
      Schluter, Natalie  and
      Tetreault, Joel",
    booktitle = "Proceedings of the 58th Annual Meeting of the Association for Computational Linguistics",
    month = jul,
    year = "2020",
    address = "Online",
    publisher = "Association for Computational Linguistics",
    url = "https://aclanthology.org/2020.acl-main.560/",
    doi = "10.18653/v1/2020.acl-main.560",
    pages = "6282--6293"
}

@article{liu-etal-2025-culturally,
    title = "Culturally Aware and Adapted {NLP}: A Taxonomy and a Survey of the State of the Art",
    author = "Liu, Chen Cecilia  and
      Gurevych, Iryna  and
      Korhonen, Anna",
    journal = "Transactions of the Association for Computational Linguistics",
    volume = "13",
    year = "2025",
    address = "Cambridge, MA",
    publisher = "MIT Press",
    url = "https://aclanthology.org/2025.tacl-1.31/",
    doi = "10.1162/tacl_a_00760",
    pages = "652--689"
}

@inproceedings{liu-etal-2025-cultural,
    title = "Cultural Learning-Based Culture Adaptation of Language Models",
    author = "Liu, Chen Cecilia  and
      Korhonen, Anna  and
      Gurevych, Iryna",
    editor = "Che, Wanxiang  and
      Nabende, Joyce  and
      Shutova, Ekaterina  and
      Pilehvar, Mohammad Taher",
    booktitle = "Proceedings of the 63rd Annual Meeting of the Association for Computational Linguistics (Volume 1: Long Papers)",
    month = jul,
    year = "2025",
    address = "Vienna, Austria",
    publisher = "Association for Computational Linguistics",
    url = "https://aclanthology.org/2025.acl-long.156/",
    doi = "10.18653/v1/2025.acl-long.156",
    pages = "3114--3134",
    ISBN = "979-8-89176-251-0"
}

@misc{nllb-team-2022-nllb,
      title={No Language Left Behind: Scaling Human-Centered Machine Translation}, 
      author={NLLB Team and Marta R. Costa-jussà and James Cross and Onur Çelebi and Maha Elbayad and Kenneth Heafield and Kevin Heffernan and Elahe Kalbassi and Janice Lam and Daniel Licht and Jean Maillard and Anna Sun and Skyler Wang and Guillaume Wenzek and Al Youngblood and Bapi Akula and Loic Barrault and Gabriel Mejia Gonzalez and Prangthip Hansanti and John Hoffman and Semarley Jarrett and Kaushik Ram Sadagopan and Dirk Rowe and Shannon Spruit and Chau Tran and Pierre Andrews and Necip Fazil Ayan and Shruti Bhosale and Sergey Edunov and Angela Fan and Cynthia Gao and Vedanuj Goswami and Francisco Guzmán and Philipp Koehn and Alexandre Mourachko and Christophe Ropers and Safiyyah Saleem and Holger Schwenk and Jeff Wang},
      year={2022},
      eprint={2207.04672},
      archivePrefix={arXiv},
      primaryClass={cs.CL},
      url={https://arxiv.org/abs/2207.04672}, 
}

@inproceedings{nyandwi-etal-2025-grounding,
    title = "Grounding Multilingual Multimodal {LLM}s With Cultural Knowledge",
    author = "Nyandwi, Jean De Dieu  and
      Song, Yueqi  and
      Khanuja, Simran  and
      Neubig, Graham",
    editor = "Christodoulopoulos, Christos  and
      Chakraborty, Tanmoy  and
      Rose, Carolyn  and
      Peng, Violet",
    booktitle = "Proceedings of the 2025 Conference on Empirical Methods in Natural Language Processing",
    month = nov,
    year = "2025",
    address = "Suzhou, China",
    publisher = "Association for Computational Linguistics",
    url = "https://aclanthology.org/2025.emnlp-main.1232/",
    doi = "10.18653/v1/2025.emnlp-main.1232",
    pages = "24187--24231",
    ISBN = "979-8-89176-332-6"
}

@inproceedings{ochieng-etal-2025-beyond,
    title = "Beyond Metrics: Evaluating {LLM}s Effectiveness in Culturally Nuanced, Low-Resource Real-World Scenarios",
    author = "Ochieng, Millicent  and
      Gumma, Varun  and
      Sitaram, Sunayana  and
      Wang, Jindong  and
      Chaudhary, Vishrav  and
      Ronen, Keshet  and
      Bali, Kalika  and
      O{'}Neill, Jacki",
    editor = "Lignos, Constantine  and
      Abdulmumin, Idris  and
      Adelani, David",
    booktitle = "Proceedings of the Sixth Workshop on African Natural Language Processing (AfricaNLP 2025)",
    month = jul,
    year = "2025",
    address = "Vienna, Austria",
    publisher = "Association for Computational Linguistics",
    url = "https://aclanthology.org/2025.africanlp-1.33/",
    doi = "10.18653/v1/2025.africanlp-1.33",
    pages = "230--247",
    ISBN = "979-8-89176-257-2"
}

@article{pawar-etal-2025-survey,
    title = "Survey of Cultural Awareness in Language Models: Text and Beyond",
    author = "Pawar, Siddhesh  and
      Park, Junyeong  and
      Jin, Jiho  and
      Arora, Arnav  and
      Myung, Junho  and
      Yadav, Srishti  and
      Haznitrama, Faiz Ghifari  and
      Song, Inhwa  and
      Oh, Alice  and
      Augenstein, Isabelle",
    journal = "Computational Linguistics",
    volume = "51",
    number = "3",
    month = sep,
    year = "2025",
    address = "Cambridge, MA",
    publisher = "MIT Press",
    url = "https://aclanthology.org/2025.cl-3.7/",
    doi = "10.1162/coli.a.14",
    pages = "907--1004"
}

@inproceedings{philippy-etal-2023-towards,
    title = "Towards a Common Understanding of Contributing Factors for Cross-Lingual Transfer in Multilingual Language Models: A Review",
    author = "Philippy, Fred  and
      Guo, Siwen  and
      Haddadan, Shohreh",
    editor = "Rogers, Anna  and
      Boyd-Graber, Jordan  and
      Okazaki, Naoaki",
    booktitle = "Proceedings of the 61st Annual Meeting of the Association for Computational Linguistics (Volume 1: Long Papers)",
    month = jul,
    year = "2023",
    address = "Toronto, Canada",
    publisher = "Association for Computational Linguistics",
    url = "https://aclanthology.org/2023.acl-long.323/",
    doi = "10.18653/v1/2023.acl-long.323",
    pages = "5877--5891"
}

@inproceedings{pranida-etal-2025-culturally,
    title = "Culturally-Nuanced Story Generation for Reasoning in Low-Resource Languages: The Case of {J}avanese and {S}undanese",
    author = "Pranida, Salsabila Zahirah  and
      Genadi, Rifo Ahmad  and
      Koto, Fajri",
    editor = "Adelani, David Ifeoluwa  and
      Arnett, Catherine  and
      Ataman, Duygu  and
      Chang, Tyler A.  and
      Gonen, Hila  and
      Raja, Rahul  and
      Schmidt, Fabian  and
      Stap, David  and
      Wang, Jiayi",
    booktitle = "Proceedings of the 5th Workshop on Multilingual Representation Learning (MRL 2025)",
    month = nov,
    year = "2025",
    address = "Suzhuo, China",
    publisher = "Association for Computational Linguistics",
    url = "https://aclanthology.org/2025.mrl-main.25/",
    doi = "10.18653/v1/2025.mrl-main.25",
    pages = "369--384",
    ISBN = "979-8-89176-345-6"
}

@inproceedings{rystrom-etal-2025-multilingual,
    title = "Multilingual != Multicultural: Evaluating Gaps Between Multilingual Capabilities and Cultural Alignment in {LLM}s",
    author = "Rystr{\o}m, Jonathan Hvithamar  and
      Kirk, Hannah Rose  and
      Hale, Scott",
    editor = "Przyby{\l}a, Piotr  and
      Shardlow, Matthew  and
      Colombatto, Clara  and
      Inie, Nanna",
    booktitle = "Proceedings of Interdisciplinary Workshop on Observations of Misunderstood, Misguided and Malicious Use of Language Models",
    month = sep,
    year = "2025",
    address = "Varna, Bulgaria",
    publisher = "INCOMA Ltd., Shoumen, Bulgaria",
    url = "https://aclanthology.org/2025.ommm-1.9/",
    pages = "74--85"
}

@inproceedings{singh-etal-2025-global,
    title = "Global {MMLU}: Understanding and Addressing Cultural and Linguistic Biases in Multilingual Evaluation",
    author = "Singh, Shivalika  and
      Romanou, Angelika  and
      Fourrier, Cl{\'e}mentine  and
      Adelani, David Ifeoluwa  and
      Ngui, Jian Gang  and
      Vila-Suero, Daniel  and
      Limkonchotiwat, Peerat  and
      Marchisio, Kelly  and
      Leong, Wei Qi  and
      Susanto, Yosephine  and
      Ng, Raymond  and
      Longpre, Shayne  and
      Ruder, Sebastian  and
      Ko, Wei-Yin  and
      Bosselut, Antoine  and
      Oh, Alice  and
      Martins, Andre  and
      Choshen, Leshem  and
      Ippolito, Daphne  and
      Ferrante, Enzo  and
      Fadaee, Marzieh  and
      Ermis, Beyza  and
      Hooker, Sara",
    editor = "Che, Wanxiang  and
      Nabende, Joyce  and
      Shutova, Ekaterina  and
      Pilehvar, Mohammad Taher",
    booktitle = "Proceedings of the 63rd Annual Meeting of the Association for Computational Linguistics (Volume 1: Long Papers)",
    month = jul,
    year = "2025",
    address = "Vienna, Austria",
    publisher = "Association for Computational Linguistics",
    url = "https://aclanthology.org/2025.acl-long.919/",
    doi = "10.18653/v1/2025.acl-long.919",
    pages = "18761--18799",
    ISBN = "979-8-89176-251-0"
}

@inproceedings{winata-etal-2025-worldcuisines,
    title = "{W}orld{C}uisines: A Massive-Scale Benchmark for Multilingual and Multicultural Visual Question Answering on Global Cuisines",
    author = "Winata, Genta Indra  and
      Hudi, Frederikus  and
      Irawan, Patrick Amadeus  and
      Anugraha, David  and
      Putri, Rifki Afina  and
      Yutong, Wang  and
      Nohejl, Adam  and
      Prathama, Ubaidillah Ariq  and
      Ousidhoum, Nedjma  and
      Amriani, Afifa  and
      Rzayev, Anar  and
      Das, Anirban  and
      Pramodya, Ashmari  and
      Adila, Aulia  and
      Wilie, Bryan  and
      Mawalim, Candy Olivia  and
      Lam, Cheng Ching  and
      Abolade, Daud  and
      Chersoni, Emmanuele  and
      Santus, Enrico  and
      Ikhwantri, Fariz  and
      Kuwanto, Garry  and
      Zhao, Hanyang  and
      Wibowo, Haryo Akbarianto  and
      Lovenia, Holy  and
      Cruz, Jan Christian Blaise  and
      Putra, Jan Wira Gotama  and
      Myung, Junho  and
      Susanto, Lucky  and
      Machin, Maria Angelica Riera  and
      Zhukova, Marina  and
      Anugraha, Michael  and
      Adilazuarda, Muhammad Farid  and
      Santosa, Natasha Christabelle  and
      Limkonchotiwat, Peerat  and
      Dabre, Raj  and
      Audino, Rio Alexander  and
      Cahyawijaya, Samuel  and
      Zhang, Shi-Xiong  and
      Salim, Stephanie Yulia  and
      Zhou, Yi  and
      Gui, Yinxuan  and
      Adelani, David Ifeoluwa  and
      Lee, En-Shiun Annie  and
      Okada, Shogo  and
      Purwarianti, Ayu  and
      Aji, Alham Fikri  and
      Watanabe, Taro  and
      Wijaya, Derry Tanti  and
      Oh, Alice  and
      Ngo, Chong-Wah",
    editor = "Chiruzzo, Luis  and
      Ritter, Alan  and
      Wang, Lu",
    booktitle = "Proceedings of the 2025 Conference of the Nations of the Americas Chapter of the Association for Computational Linguistics: Human Language Technologies (Volume 1: Long Papers)",
    month = apr,
    year = "2025",
    address = "Albuquerque, New Mexico",
    publisher = "Association for Computational Linguistics",
    url = "https://aclanthology.org/2025.naacl-long.167/",
    doi = "10.18653/v1/2025.naacl-long.167",
    pages = "3242--3264",
    ISBN = "979-8-89176-189-6"
}

@inproceedings{limisiewicz-etal-2023-tokenization,
    title = "Tokenization Impacts Multilingual Language Modeling: Assessing Vocabulary Allocation and Overlap Across Languages",
    author = "Limisiewicz, Tomasz  and
      Balhar, Ji{\v{r}}{\'i}  and
      Mare{\v{c}}ek, David",
    editor = "Rogers, Anna  and
      Boyd-Graber, Jordan  and
      Okazaki, Naoaki",
    booktitle = "Findings of the Association for Computational Linguistics: ACL 2023",
    month = jul,
    year = "2023",
    address = "Toronto, Canada",
    publisher = "Association for Computational Linguistics",
    url = "https://aclanthology.org/2023.findings-acl.350/",
    doi = "10.18653/v1/2023.findings-acl.350",
    pages = "5661--5681"
}

@inproceedings{alkhamissi-etal-2024-investigating,
    title = "Investigating Cultural Alignment of Large Language Models",
    author = "AlKhamissi, Badr  and
      ElNokrashy, Muhammad  and
      Alkhamissi, Mai  and
      Diab, Mona",
    editor = "Ku, Lun-Wei  and
      Martins, Andre  and
      Srikumar, Vivek",
    booktitle = "Proceedings of the 62nd Annual Meeting of the Association for Computational Linguistics (Volume 1: Long Papers)",
    month = aug,
    year = "2024",
    address = "Bangkok, Thailand",
    publisher = "Association for Computational Linguistics",
    url = "https://aclanthology.org/2024.acl-long.671/",
    doi = "10.18653/v1/2024.acl-long.671",
    pages = "12404--12422"
}

@inproceedings{naous-etal-2024-beer,
    title = "Having Beer after Prayer? Measuring Cultural Bias in Large Language Models",
    author = "Naous, Tarek  and
      Ryan, Michael J  and
      Ritter, Alan  and
      Xu, Wei",
    editor = "Ku, Lun-Wei  and
      Martins, Andre  and
      Srikumar, Vivek",
    booktitle = "Proceedings of the 62nd Annual Meeting of the Association for Computational Linguistics (Volume 1: Long Papers)",
    month = aug,
    year = "2024",
    address = "Bangkok, Thailand",
    publisher = "Association for Computational Linguistics",
    url = "https://aclanthology.org/2024.acl-long.862/",
    doi = "10.18653/v1/2024.acl-long.862",
    pages = "16366--16393"
}

@inproceedings{wang-etal-2024-cdeval,
    title = "{CDE}val: A Benchmark for Measuring the Cultural Dimensions of Large Language Models",
    author = "Wang, Yuhang  and
      Zhu, Yanxu  and
      Kong, Chao  and
      Wei, Shuyu  and
      Yi, Xiaoyuan  and
      Xie, Xing  and
      Sang, Jitao",
    editor = "Prabhakaran, Vinodkumar  and
      Dev, Sunipa  and
      Benotti, Luciana  and
      Hershcovich, Daniel  and
      Cabello, Laura  and
      Cao, Yong  and
      Adebara, Ife  and
      Zhou, Li",
    booktitle = "Proceedings of the 2nd Workshop on Cross-Cultural Considerations in NLP",
    month = aug,
    year = "2024",
    address = "Bangkok, Thailand",
    publisher = "Association for Computational Linguistics",
    url = "https://aclanthology.org/2024.c3nlp-1.1/",
    doi = "10.18653/v1/2024.c3nlp-1.1",
    pages = "1--16"
}

@inproceedings{nayak-etal-2024-benchmarking,
    title = "Benchmarking Vision Language Models for Cultural Understanding",
    author = "Nayak, Shravan  and
      Jain, Kanishk  and
      Awal, Rabiul  and
      Reddy, Siva  and
      Steenkiste, Sjoerd Van  and
      Hendricks, Lisa Anne  and
      Stanczak, Karolina  and
      Agrawal, Aishwarya",
    editor = "Al-Onaizan, Yaser  and
      Bansal, Mohit  and
      Chen, Yun-Nung",
    booktitle = "Proceedings of the 2024 Conference on Empirical Methods in Natural Language Processing",
    month = nov,
    year = "2024",
    address = "Miami, Florida, USA",
    publisher = "Association for Computational Linguistics",
    url = "https://aclanthology.org/2024.emnlp-main.329/",
    doi = "10.18653/v1/2024.emnlp-main.329",
    pages = "5769--5790"
}

@inproceedings{adilazuarda-etal-2024-towards,
    title = "Towards Measuring and Modeling ``Culture'' in {LLM}s: A Survey",
    author = "Adilazuarda, Muhammad Farid  and
      Mukherjee, Sagnik  and
      Lavania, Pradhyumna  and
      Singh, Siddhant Shivdutt  and
      Aji, Alham Fikri  and
      O{'}Neill, Jacki  and
      Modi, Ashutosh  and
      Choudhury, Monojit",
    editor = "Al-Onaizan, Yaser  and
      Bansal, Mohit  and
      Chen, Yun-Nung",
    booktitle = "Proceedings of the 2024 Conference on Empirical Methods in Natural Language Processing",
    month = nov,
    year = "2024",
    address = "Miami, Florida, USA",
    publisher = "Association for Computational Linguistics",
    url = "https://aclanthology.org/2024.emnlp-main.882/",
    doi = "10.18653/v1/2024.emnlp-main.882",
    pages = "15763--15784"
}

@inproceedings{zhao-etal-2024-worldvaluesbench,
    title = "{W}orld{V}alues{B}ench: A Large-Scale Benchmark Dataset for Multi-Cultural Value Awareness of Language Models",
    author = "Zhao, Wenlong  and
      Mondal, Debanjan  and
      Tandon, Niket  and
      Dillion, Danica  and
      Gray, Kurt  and
      Gu, Yuling",
    editor = "Calzolari, Nicoletta  and
      Kan, Min-Yen  and
      Hoste, Veronique  and
      Lenci, Alessandro  and
      Sakti, Sakriani  and
      Xue, Nianwen",
    booktitle = "Proceedings of the 2024 Joint International Conference on Computational Linguistics, Language Resources and Evaluation (LREC-COLING 2024)",
    month = may,
    year = "2024",
    address = "Torino, Italia",
    publisher = "ELRA and ICCL",
    url = "https://aclanthology.org/2024.lrec-main.1539/",
    pages = "17696--17706"
}

@inproceedings{chiu-etal-2025-culturalbench,
    title = "{C}ultural{B}ench: A Robust, Diverse and Challenging Benchmark for Measuring {LM}s' Cultural Knowledge Through Human-{AI} Red-Teaming",
    author = "Chiu, Yu Ying  and
      Jiang, Liwei  and
      Lin, Bill Yuchen  and
      Park, Chan Young  and
      Li, Shuyue Stella  and
      Ravi, Sahithya  and
      Bhatia, Mehar  and
      Antoniak, Maria  and
      Tsvetkov, Yulia  and
      Shwartz, Vered  and
      Choi, Yejin",
    editor = "Che, Wanxiang  and
      Nabende, Joyce  and
      Shutova, Ekaterina  and
      Pilehvar, Mohammad Taher",
    booktitle = "Proceedings of the 63rd Annual Meeting of the Association for Computational Linguistics (Volume 1: Long Papers)",
    month = jul,
    year = "2025",
    address = "Vienna, Austria",
    publisher = "Association for Computational Linguistics",
    url = "https://aclanthology.org/2025.acl-long.1247/",
    doi = "10.18653/v1/2025.acl-long.1247",
    pages = "25663--25701",
    ISBN = "979-8-89176-251-0"
}

@inproceedings{zhou-etal-2025-culture,
    title = "Culture is Not Trivia: Sociocultural Theory for Cultural {NLP}",
    author = "Zhou, Naitian  and
      Bamman, David  and
      Bleaman, Isaac L.",
    editor = "Che, Wanxiang  and
      Nabende, Joyce  and
      Shutova, Ekaterina  and
      Pilehvar, Mohammad Taher",
    booktitle = "Proceedings of the 63rd Annual Meeting of the Association for Computational Linguistics (Volume 1: Long Papers)",
    month = jul,
    year = "2025",
    address = "Vienna, Austria",
    publisher = "Association for Computational Linguistics",
    url = "https://aclanthology.org/2025.acl-long.1256/",
    doi = "10.18653/v1/2025.acl-long.1256",
    pages = "25869--25886",
    ISBN = "979-8-89176-251-0"
}

@inproceedings{yang-etal-2025-cultural,
    title = "Cultural Bias Matters: A Cross-Cultural Benchmark Dataset and Sentiment-Enriched Model for Understanding Multimodal Metaphors",
    author = "Yang, Senqi  and
      Zhang, Dongyu  and
      Ren, Jing  and
      Xu, Ziqi  and
      Zhang, Xiuzhen  and
      Song, Yiliao  and
      Lin, Hongfei  and
      Xia, Feng",
    editor = "Che, Wanxiang  and
      Nabende, Joyce  and
      Shutova, Ekaterina  and
      Pilehvar, Mohammad Taher",
    booktitle = "Proceedings of the 63rd Annual Meeting of the Association for Computational Linguistics (Volume 1: Long Papers)",
    month = jul,
    year = "2025",
    address = "Vienna, Austria",
    publisher = "Association for Computational Linguistics",
    url = "https://aclanthology.org/2025.acl-long.1275/",
    doi = "10.18653/v1/2025.acl-long.1275",
    pages = "26301--26317",
    ISBN = "979-8-89176-251-0"
}

@inproceedings{masoud-etal-2025-cultural,
    title = "Cultural Alignment in Large Language Models: An Explanatory Analysis Based on Hofstede{'}s Cultural Dimensions",
    author = "Masoud, Reem I.  and
      Liu, Ziquan  and
      Ferianc, Martin  and
      Treleaven, Philip  and
      Rodrigues, Miguel",
    editor = "Rambow, Owen  and
      Wanner, Leo  and
      Apidianaki, Marianna  and
      Al-Khalifa, Hend  and
      Eugenio, Barbara Di  and
      Schockaert, Steven",
    booktitle = "Proceedings of the 31st International Conference on Computational Linguistics",
    month = jan,
    year = "2025",
    address = "Abu Dhabi, UAE",
    publisher = "Association for Computational Linguistics",
    url = "https://aclanthology.org/2025.coling-main.567/",
    pages = "8474--8503"
}

@inproceedings{maji-etal-2025-drishtikon,
    title = "{DRISHTIKON}: A Multimodal Multilingual Benchmark for Testing Language Models' Understanding on {I}ndian Culture",
    author = "Maji, Arijit  and
      Kumar, Raghvendra  and
      Ghosh, Akash  and
      Anushka  and
      Shah, Nemil  and
      Borah, Abhilekh  and
      Shah, Vanshika  and
      Mishra, Nishant  and
      Saha, Sriparna",
    editor = "Christodoulopoulos, Christos  and
      Chakraborty, Tanmoy  and
      Rose, Carolyn  and
      Peng, Violet",
    booktitle = "Proceedings of the 2025 Conference on Empirical Methods in Natural Language Processing",
    month = nov,
    year = "2025",
    address = "Suzhou, China",
    publisher = "Association for Computational Linguistics",
    url = "https://aclanthology.org/2025.emnlp-main.68/",
    doi = "10.18653/v1/2025.emnlp-main.68",
    pages = "1289--1313",
    ISBN = "979-8-89176-332-6"
}

@inproceedings{aji-cohn-2025-loraxbench,
    title = "{LORAXBENCH}: A Multitask, Multilingual Benchmark Suite for 20 {I}ndonesian Languages",
    author = "Aji, Alham Fikri  and
      Cohn, Trevor",
    editor = "Christodoulopoulos, Christos  and
      Chakraborty, Tanmoy  and
      Rose, Carolyn  and
      Peng, Violet",
    booktitle = "Proceedings of the 2025 Conference on Empirical Methods in Natural Language Processing",
    month = nov,
    year = "2025",
    address = "Suzhou, China",
    publisher = "Association for Computational Linguistics",
    url = "https://aclanthology.org/2025.emnlp-main.881/",
    doi = "10.18653/v1/2025.emnlp-main.881",
    pages = "17421--17446",
    ISBN = "979-8-89176-332-6"
}

@inproceedings{schneider-etal-2025-gimmick,
    title = "{GIMMICK}: Globally Inclusive Multimodal Multitask Cultural Knowledge Benchmarking",
    author = "Schneider, Florian  and
      Holtermann, Carolin  and
      Biemann, Chris  and
      Lauscher, Anne",
    editor = "Che, Wanxiang  and
      Nabende, Joyce  and
      Shutova, Ekaterina  and
      Pilehvar, Mohammad Taher",
    booktitle = "Findings of the Association for Computational Linguistics: ACL 2025",
    month = jul,
    year = "2025",
    address = "Vienna, Austria",
    publisher = "Association for Computational Linguistics",
    url = "https://aclanthology.org/2025.findings-acl.500/",
    doi = "10.18653/v1/2025.findings-acl.500",
    pages = "9605--9668",
    ISBN = "979-8-89176-256-5"
}

@inproceedings{kim-lee-2025-nunchi,
    title = "Nunchi-Bench: Benchmarking Language Models on Cultural Reasoning with a Focus on {K}orean Superstition",
    author = "Kim, Kyuhee  and
      Lee, Sangah",
    editor = "Che, Wanxiang  and
      Nabende, Joyce  and
      Shutova, Ekaterina  and
      Pilehvar, Mohammad Taher",
    booktitle = "Findings of the Association for Computational Linguistics: ACL 2025",
    month = jul,
    year = "2025",
    address = "Vienna, Austria",
    publisher = "Association for Computational Linguistics",
    url = "https://aclanthology.org/2025.findings-acl.794/",
    doi = "10.18653/v1/2025.findings-acl.794",
    pages = "15328--15342",
    ISBN = "979-8-89176-256-5"
}

@inproceedings{villa-cueva-etal-2025-cammt,
    title = "{C}a{MMT}: Benchmarking Culturally Aware Multimodal Machine Translation",
    author = "Villa-Cueva, Emilio  and
      Bolatzhanova, Sholpan  and
      Turmakhan, Diana  and
      Elzeky, Kareem  and
      Ademtew, Henok Biadglign  and
      Aji, Alham Fikri  and
      Araujo, Vladimir  and
      Azime, Israel Abebe  and
      Baek, Jinheon  and
      Belcavello, Frederico  and
      Cristobal, Fermin  and
      Cruz, Jan Christian Blaise  and
      Dabre, Mary  and
      Dabre, Raj  and
      Ehsan, Toqeer  and
      Etori, Naome A  and
      Farooqui, Fauzan  and
      Geng, Jiahui  and
      Ivetta, Guido  and
      Jayakumar, Thanmay  and
      Jeong, Soyeong  and
      Lim, Zheng Wei  and
      Mandal, Aishik  and
      Martinelli, Sof{\'i}a  and
      Mihaylov, Mihail Minkov  and
      Orel, Daniil  and
      Pramanick, Aniket  and
      Purkayastha, Sukannya  and
      Salazar, Israfel  and
      Song, Haiyue  and
      Timponi Torrent, Tiago  and
      Yadeta, Debela Desalegn  and
      Hamed, Injy  and
      Tonja, Atnafu Lambebo  and
      Solorio, Thamar",
    editor = "Christodoulopoulos, Christos  and
      Chakraborty, Tanmoy  and
      Rose, Carolyn  and
      Peng, Violet",
    booktitle = "Findings of the Association for Computational Linguistics: EMNLP 2025",
    month = nov,
    year = "2025",
    address = "Suzhou, China",
    publisher = "Association for Computational Linguistics",
    url = "https://aclanthology.org/2025.findings-emnlp.1220/",
    doi = "10.18653/v1/2025.findings-emnlp.1220",
    pages = "22423--22441",
    ISBN = "979-8-89176-335-7"
}

@inproceedings{havaldar-etal-2025-culturally,
    title = "Culturally-Aware Conversations: A Framework {\&} Benchmark for {LLM}s",
    author = "Havaldar, Shreya  and
      Cho, Young Min  and
      Rai, Sunny  and
      Ungar, Lyle",
    editor = "Blodgett, Su Lin  and
      Curry, Amanda Cercas  and
      Dev, Sunipa  and
      Li, Siyan  and
      Madaio, Michael  and
      Wang, Jack  and
      Wu, Sherry Tongshuang  and
      Xiao, Ziang  and
      Yang, Diyi",
    booktitle = "Proceedings of the Fourth Workshop on Bridging Human-Computer Interaction and Natural Language Processing (HCI+NLP)",
    month = nov,
    year = "2025",
    address = "Suzhou, China",
    publisher = "Association for Computational Linguistics",
    url = "https://aclanthology.org/2025.hcinlp-1.18/",
    doi = "10.18653/v1/2025.hcinlp-1.18",
    pages = "220--229",
    ISBN = "979-8-89176-353-1"
}

@inproceedings{naous-xu-2025-origin,
    title = "On The Origin of Cultural Biases in Language Models: From Pre-training Data to Linguistic Phenomena",
    author = "Naous, Tarek  and
      Xu, Wei",
    editor = "Chiruzzo, Luis  and
      Ritter, Alan  and
      Wang, Lu",
    booktitle = "Proceedings of the 2025 Conference of the Nations of the Americas Chapter of the Association for Computational Linguistics: Human Language Technologies (Volume 1: Long Papers)",
    month = apr,
    year = "2025",
    address = "Albuquerque, New Mexico",
    publisher = "Association for Computational Linguistics",
    url = "https://aclanthology.org/2025.naacl-long.326/",
    doi = "10.18653/v1/2025.naacl-long.326",
    pages = "6423--6443",
    ISBN = "979-8-89176-189-6"
}

@inproceedings{xu-etal-2025-self,
    title = "Self-Pluralising Culture Alignment for Large Language Models",
    author = "Xu, Shaoyang  and
      Leng, Yongqi  and
      Yu, Linhao  and
      Xiong, Deyi",
    editor = "Chiruzzo, Luis  and
      Ritter, Alan  and
      Wang, Lu",
    booktitle = "Proceedings of the 2025 Conference of the Nations of the Americas Chapter of the Association for Computational Linguistics: Human Language Technologies (Volume 1: Long Papers)",
    month = apr,
    year = "2025",
    address = "Albuquerque, New Mexico",
    publisher = "Association for Computational Linguistics",
    url = "https://aclanthology.org/2025.naacl-long.350/",
    doi = "10.18653/v1/2025.naacl-long.350",
    pages = "6859--6877",
    ISBN = "979-8-89176-189-6"
}

@inproceedings{hammerl-etal-2025-beyond,
    title = "Beyond Literal Token Overlap: Token Alignability for Multilinguality",
    author = {H{\"a}mmerl, Katharina  and
      Limisiewicz, Tomasz  and
      Libovick{\'y}, Jind{\v{r}}ich  and
      Fraser, Alexander},
    editor = "Chiruzzo, Luis  and
      Ritter, Alan  and
      Wang, Lu",
    booktitle = "Proceedings of the 2025 Conference of the Nations of the Americas Chapter of the Association for Computational Linguistics: Human Language Technologies (Volume 2: Short Papers)",
    month = apr,
    year = "2025",
    address = "Albuquerque, New Mexico",
    publisher = "Association for Computational Linguistics",
    url = "https://aclanthology.org/2025.naacl-short.63/",
    doi = "10.18653/v1/2025.naacl-short.63",
    pages = "756--767",
    ISBN = "979-8-89176-190-2"
}

@inproceedings{oh-etal-2025-culture,
    title = "Culture is Everywhere: A Call for Intentionally Cultural Evaluation",
    author = "Oh, Juhyun  and
      Cha, Inha  and
      Saxon, Michael  and
      Lim, Hyunseung  and
      Bhatt, Shaily  and
      Oh, Alice",
    editor = "Christodoulopoulos, Christos  and
      Chakraborty, Tanmoy  and
      Rose, Carolyn  and
      Peng, Violet",
    booktitle = "Findings of the Association for Computational Linguistics: EMNLP 2025",
    month = nov,
    year = "2025",
    address = "Suzhou, China",
    publisher = "Association for Computational Linguistics",
    url = "https://aclanthology.org/2025.findings-emnlp.1043/",
    doi = "10.18653/v1/2025.findings-emnlp.1043",
    pages = "19156--19168",
    ISBN = "979-8-89176-335-7"
}

@inproceedings{wu-etal-2025-incorporating,
    title = "Incorporating Diverse Perspectives in Cultural Alignment: Survey of Evaluation Benchmarks Through A Three-Dimensional Framework",
    author = "Wu, Meng-Chen  and
      Chin, Si-Chi  and
      Wood, Tess  and
      Goyal, Ayush  and
      Sadagopan, Narayanan",
    editor = "Christodoulopoulos, Christos  and
      Chakraborty, Tanmoy  and
      Rose, Carolyn  and
      Peng, Violet",
    booktitle = "Proceedings of the 2025 Conference on Empirical Methods in Natural Language Processing",
    month = nov,
    year = "2025",
    address = "Suzhou, China",
    publisher = "Association for Computational Linguistics",
    url = "https://aclanthology.org/2025.emnlp-main.862/",
    doi = "10.18653/v1/2025.emnlp-main.862",
    pages = "17026--17061",
    ISBN = "979-8-89176-332-6"
}

@inproceedings{qiu-etal-2025-evaluating,
    title = "Evaluating Cultural and Social Awareness of {LLM} Web Agents",
    author = "Qiu, Haoyi  and
      Fabbri, Alexander  and
      Agarwal, Divyansh  and
      Huang, Kung-Hsiang  and
      Tan, Sarah  and
      Peng, Nanyun  and
      Wu, Chien-Sheng",
    editor = "Chiruzzo, Luis  and
      Ritter, Alan  and
      Wang, Lu",
    booktitle = "Findings of the Association for Computational Linguistics: NAACL 2025",
    month = apr,
    year = "2025",
    address = "Albuquerque, New Mexico",
    publisher = "Association for Computational Linguistics",
    url = "https://aclanthology.org/2025.findings-naacl.222/",
    doi = "10.18653/v1/2025.findings-naacl.222",
    pages = "3978--4005",
    ISBN = "979-8-89176-195-7"
}

@inproceedings{muller-eberstein-etal-2025-dakultur,
    title = "{D}a{K}ultur: Evaluating the Cultural Awareness of Language Models for {D}anish with Native Speakers",
    author = {M{\"u}ller-Eberstein, Max  and
      Zhang, Mike  and
      Bassignana, Elisa  and
      Trolle, Peter Brunsgaard  and
      Goot, Rob Van Der},
    editor = "Prabhakaran, Vinodkumar  and
      Dev, Sunipa  and
      Benotti, Luciana  and
      Hershcovich, Daniel  and
      Cao, Yong  and
      Zhou, Li  and
      Cabello, Laura  and
      Adebara, Ife",
    booktitle = "Proceedings of the 3rd Workshop on Cross-Cultural Considerations in NLP (C3NLP 2025)",
    month = may,
    year = "2025",
    address = "Albuquerque, New Mexico",
    publisher = "Association for Computational Linguistics",
    url = "https://aclanthology.org/2025.c3nlp-1.5/",
    doi = "10.18653/v1/2025.c3nlp-1.5",
    pages = "50--58",
    ISBN = "979-8-89176-237-4"
}

@inproceedings{pham-etal-2025-cultureinstruct,
    title = "{C}ulture{I}nstruct: Curating Multi-Cultural Instructions at Scale",
    author = "Pham, Viet Thanh  and
      Li, Zhuang  and
      Qu, Lizhen  and
      Haffari, Gholamreza",
    editor = "Chiruzzo, Luis  and
      Ritter, Alan  and
      Wang, Lu",
    booktitle = "Proceedings of the 2025 Conference of the Nations of the Americas Chapter of the Association for Computational Linguistics: Human Language Technologies (Volume 1: Long Papers)",
    month = apr,
    year = "2025",
    address = "Albuquerque, New Mexico",
    publisher = "Association for Computational Linguistics",
    url = "https://aclanthology.org/2025.naacl-long.465/",
    doi = "10.18653/v1/2025.naacl-long.465",
    pages = "9207--9228",
    ISBN = "979-8-89176-189-6"
}

@inproceedings{kabir-etal-2025-break,
    title = "Break the Checkbox: Challenging Closed-Style Evaluations of Cultural Alignment in {LLM}s",
    author = "Kabir, Mohsinul  and
      Abrar, Ajwad  and
      Ananiadou, Sophia",
    editor = "Christodoulopoulos, Christos  and
      Chakraborty, Tanmoy  and
      Rose, Carolyn  and
      Peng, Violet",
    booktitle = "Proceedings of the 2025 Conference on Empirical Methods in Natural Language Processing",
    month = nov,
    year = "2025",
    address = "Suzhou, China",
    publisher = "Association for Computational Linguistics",
    url = "https://aclanthology.org/2025.emnlp-main.2/",
    doi = "10.18653/v1/2025.emnlp-main.2",
    pages = "24--51",
    ISBN = "979-8-89176-332-6"
}

@inproceedings{faisal-etal-2024-dialectbench,
    title = "{DIALECTBENCH}: An {NLP} Benchmark for Dialects, Varieties, and Closely-Related Languages",
    author = "Faisal, Fahim  and
      Ahia, Orevaoghene  and
      Srivastava, Aarohi  and
      Ahuja, Kabir  and
      Chiang, David  and
      Tsvetkov, Yulia  and
      Anastasopoulos, Antonios",
    editor = "Ku, Lun-Wei  and
      Martins, Andre  and
      Srikumar, Vivek",
    booktitle = "Proceedings of the 62nd Annual Meeting of the Association for Computational Linguistics (Volume 1: Long Papers)",
    month = aug,
    year = "2024",
    address = "Bangkok, Thailand",
    publisher = "Association for Computational Linguistics",
    url = "https://aclanthology.org/2024.acl-long.777/",
    doi = "10.18653/v1/2024.acl-long.777",
    pages = "14412--14454"
}

@inproceedings{shafique-etal-2025-culturally,
    title = "A Culturally-diverse Multilingual Multimodal Video Benchmark {\&} Model",
    author = "Shafique, Bhuiyan Sanjid  and
      Vayani, Ashmal  and
      Maaz, Muhammad  and
      Rasheed, Hanoona Abdul  and
      Dissanayake, Dinura  and
      Kurpath, Mohammed Irfan  and
      Hmaiti, Yahya  and
      Inoue, Go  and
      Lahoud, Jean  and
      Rashid, Md. Safirur  and
      Quasem, Shadid Intisar  and
      Fatima, Maheen  and
      Vidal, Franco  and
      Maslych, Mykola  and
      More, Ketan Pravin  and
      Baliah, Sanoojan  and
      Watawana, Hasindri  and
      Li, Yuhao  and
      Farestam, Fabian  and
      Schaller, Leon  and
      Tymtsiv, Roman  and
      Weber, Simon  and
      Cholakkal, Hisham  and
      Laptev, Ivan  and
      Satoh, Shin{'}ichi  and
      Felsberg, Michael  and
      Shah, Mubarak  and
      Khan, Salman  and
      Khan, Fahad Shahbaz",
    editor = "Christodoulopoulos, Christos  and
      Chakraborty, Tanmoy  and
      Rose, Carolyn  and
      Peng, Violet",
    booktitle = "Proceedings of the 2025 Conference on Empirical Methods in Natural Language Processing",
    month = nov,
    year = "2025",
    address = "Suzhou, China",
    publisher = "Association for Computational Linguistics",
    url = "https://aclanthology.org/2025.emnlp-main.1012/",
    doi = "10.18653/v1/2025.emnlp-main.1012",
    pages = "19998--20022",
    ISBN = "979-8-89176-332-6"
}

@inproceedings{caselli-etal-2021-guiding,
    title = "Guiding Principles for Participatory Design-inspired Natural Language Processing",
    author = "Caselli, Tommaso  and
      Cibin, Roberto  and
      Conforti, Costanza  and
      Encinas, Enrique  and
      Teli, Maurizio",
    editor = "Field, Anjalie  and
      Prabhumoye, Shrimai  and
      Sap, Maarten  and
      Jin, Zhijing  and
      Zhao, Jieyu  and
      Brockett, Chris",
    booktitle = "Proceedings of the 1st Workshop on NLP for Positive Impact",
    month = aug,
    year = "2021",
    address = "Online",
    publisher = "Association for Computational Linguistics",
    url = "https://aclanthology.org/2021.nlp4posimpact-1.4/",
    doi = "10.18653/v1/2021.nlp4posimpact-1.4",
    pages = "27--35"
}

@inproceedings{ivetta-etal-2025-heseia,
    title = "{HESEIA}: A community-based dataset for evaluating social biases in large language models, co-designed in real school settings in {L}atin {A}merica",
    author = "Ivetta, Guido  and
      Gomez, Marcos J  and
      Martinelli, Sof{\'i}a  and
      Palombini, Pietro  and
      Echeveste, M Emilia  and
      Mazzeo, Nair Carolina  and
      Busaniche, Beatriz  and
      Benotti, Luciana",
    editor = "Christodoulopoulos, Christos  and
      Chakraborty, Tanmoy  and
      Rose, Carolyn  and
      Peng, Violet",
    booktitle = "Proceedings of the 2025 Conference on Empirical Methods in Natural Language Processing",
    month = nov,
    year = "2025",
    address = "Suzhou, China",
    publisher = "Association for Computational Linguistics",
    url = "https://aclanthology.org/2025.emnlp-main.1275/",
    doi = "10.18653/v1/2025.emnlp-main.1275",
    pages = "25095--25117",
    ISBN = "979-8-89176-332-6"
}

\end{document}